\documentclass[runningheads]{llncs}
\usepackage{graphicx}

\usepackage{tikz}
\usepackage{comment}
\usepackage{amsmath,amssymb} 
\usepackage{color}

\newcommand*\samethanks[1][\value{footnote}]{\footnotemark[#1]}
\usepackage{multirow}
\usepackage{wrapfig}
\usepackage{changepage}
\usepackage{caption}
\usepackage{booktabs}
\usepackage{hyperref}
\usepackage[capitalize]{cleveref}
\crefname{section}{Sec.}{Secs.}
\Crefname{section}{Section}{Sections}
\Crefname{table}{Table}{Tables}
\crefname{table}{Tab.}{Tabs.}

\newcommand{\CLIPa}{CLIP (ViT-B/32)}
\newcommand{\CLIPb}{CLIP (ViT-B/32, T768)}
\newcommand{\MACLIPa}{MS-CLIP (B/32)}
\newcommand{\MACLIPb}{MS-CLIP (B/32) + Early Specialization}

\newcommand{\MACLIPd}{MS-CLIP-S (B/32)}

\usepackage[accsupp]{axessibility}  

\begin{document}
\pagestyle{headings}
\mainmatter
\def\ECCVSubNumber{7004}  

\title{Learning Visual Representation from Modality-Shared Contrastive Language-Image Pre-training}  

\titlerunning{MS-CLIP}
%
\author{Haoxuan You\inst{1}\thanks{Equal Contribution} \and
Luowei Zhou\inst{2}\samethanks[1] \and
Bin Xiao\inst{2}\samethanks[1] \and
Noel Codella\inst{2}\samethanks[1]  \and \\
Yu Cheng\inst{3} \and
Ruochen Xu\inst{2} \and
Shih-Fu Chang\inst{1} \and
Lu Yuan\inst{2}
}
\authorrunning{H. You et al.}
%
\institute{Columbia University, New York, USA \and
Microsoft Cloud and AI, Redmond, USA \and Microsoft Research, Redmond, USA \\
\email{\{hy2612,sc250\}@columbia.edu}\\
\email{\{luozhou,bixi,ncodella,yu.cheng,ruox,luyuan\}@microsoft.com}}
\maketitle

\begin{abstract}
Large-scale multi-modal contrastive pre-training has demonstrated great utility to learn transferable features for a range of downstream tasks by mapping multiple modalities into a shared embedding space. Typically, this has employed separate encoders for each modality. However, recent work suggests that transformers can support learning across multiple modalities and allow knowledge sharing. Inspired by this, we investigate a variety of Modality-Shared Contrastive Language-Image Pre-training (MS-CLIP) frameworks. More specifically, we question how many parameters of a transformer model can be shared across modalities during contrastive pre-training, and rigorously examine architectural design choices that position the proportion of parameters shared along a spectrum. In studied conditions, we observe that a mostly unified encoder for vision and language signals outperforms all other variations that separate more parameters. Additionally, we find that light-weight modality-specific parallel modules further improve performance. Experimental results show that the proposed MS-CLIP approach outperforms vanilla CLIP by up to 13\% relative in zero-shot ImageNet classification (pre-trained on YFCC-100M), while simultaneously supporting a reduction of parameters. In addition, our approach outperforms vanilla CLIP by 1.6 points in linear probing on a collection of 24 downstream vision tasks. Furthermore, we discover that sharing parameters leads to semantic concepts from different modalities being encoded more closely in the embedding space, facilitating the transferring of common semantic structure (e.g., attention patterns) from language to vision. Code is available at \href{https://github.com/Hxyou/MSCLIP}{https://github.com/Hxyou/MSCLIP}.
\end{abstract}

\section{Introduction}
Contrastive Language-Image Pre-training (CLIP) has drawn much attention recently in the field of Computer Vision and Natural Language Processing \cite{jia2021scaling,radford2021learning}, where large-scale image-caption data are leveraged to learn generic vision representations from language supervision through contrastive loss. This allows the learning of open-set visual concepts and imbues the learned features with a robust capability to transfer to diverse vision tasks. 

Prior work in this topic often employs separate language and image encoders, despite architectural similarities between the encoders for both modalities. For instance, the original CLIP work~\cite{radford2021learning} uses a ViT ~\cite{dosovitskiy2020image} based image encoder, and a separate transformer~\cite{vaswani2017attention} based language encoder.
However, another work~\cite{lu2021pretrained} recently discovered that transformer models pre-trained on language data could generalize well to visual tasks without altering the majority of parameters, suggesting patterns learned by one modality could transfer to another. These observations suggest that a unified encoder for CLIP may potentially be leveraged to promote learning commonly useful representations across modalities to realize performance and efficiency gains.




In this paper, we consequently investigate the feasibility of building a Modality-Shared CLIP (MS-CLIP) architecture, where parameters in vision encoder and text encoder can be shared. 
Through this framework, we seek answers to the following three questions: ($i$) Within each layer, which sub-module should be shared and which should not? ($ii$) In the CLIP training setting, which layers of the encoders for the two modalities should be shared, and which should be modality-specific? ($iii$) Lastly, what is the impact to performance and efficiency when including lightweight modality-specific auxiliary modules to accommodate specializations in each modality?

In order to answer these questions, we perform a comprehensive analysis on the impact of varying the degree of sharing of components across different layers. Our results show that in order to maximize performance, the input embedding, layer normalization (LN) \cite{ba2016layer}, and output projection should be modality-specific. In contrast, all the remaining components can be shared across vision and text transformers, including the weights in self-attention and feed-forward modules. Addtionally, sharing all transformer layers even outperforms more complex strategies where we employ greedy selection of layers or use Neural Architecture Search (NAS)~\cite{dong2019searching} to search for the optimal layer sharing policy.


Finally, we explore whether introducing lightweight modality-specific components to the shared backbone may yield a better balance between cross-modality modeling and specializations within each modality. Studied designs include: ($i$) {\em Early Specialization:} The first Transformer block is replaced by modules that are specialized for each modality, respectively. This includes a set of lightweight cascaded residual convolutional neural networks (CNNs) for vision, and a Transformer layer for language. This early adaption allows the representation in each modality to abstract to a higher level before unified encoding, and introduces shift invariance early in the visual branch.
($ii$) {\em Efficient Parallel Branch:} For the visual modality, we explore a lightweight multi-scale CNN network, parallel to the main modality-shared branch, and incorporate its multi-scale features to the main branch through depth-wise convolutional adaptors. This parallel branch enables augmenting the main branch with the benefits convolutions can instill from better modeling of spatial relationships.


We pre-train MS-CLIP architectures on YFCC100M~\cite{thomee2016yfcc100m} and a subset of Laion-400M~\cite{schuhmann2021laion} with a similar size, and evaluate on 25 downstream datasets that encompass a broad variety of vision tasks. The experimental results demonstrate that MS-CLIP architectures, while having fewer parameters, can outperform original CLIP on the majority of tasks, including zero-shot recognition, zero-shot retrieval, and linear probing. Moreover, in order to better understand why MS-CLIP architectures work so well, we conduct studies on the learned embedding space, namely with a measurement on multi-modal feature fusion degree~\cite{cao2020behind}, and quantitatively assess to what degree semantic structures (e.g., attention patterns) are shared across modalities. 
Our results reveal that sharing parameters can pull semantically-similar concepts from different modalities closer and facilitate the learning of common semantic structures (e.g., attention patterns).


The paper is subsequently organized as follows. Section \ref{sec:relatedwork} covers related work. In Section \ref{sec:methods}, we introduce the shareable modules and modality-specific designs. In Section \ref{sec:adapters}, we present a rigorous study varying amount of parameters shared across modalities and measure the impact of both modality-shared parameters and modality-specific modules to downstream performance and efficiency. And we comprehensively compare proposed MS-CLIP architectures against CLIP on 25 downstream datasets.  Section \ref{sec:conclusion} concludes. 



\section{Related Work}
\label{sec:relatedwork}
\paragraph{Learning Visual Representation from Text:}
Our work is built on the recent success of learning visual representation from text supervision.
VirTex~\cite{desai2021virtex} proposes to learn visual encoding through an image captioning objective. LocTex~\cite{liu2021loctex} introduces localized textual supervision to guide visual representation learning. Both studies are conducted on a relatively small scale. More recent work such as CLIP~\cite{radford2021learning} and ALIGN~\cite{jia2021scaling} demonstrate that generic multi-modal pre-training could benefit from extremely large scale training (i.e., private datasets with hundreds of millions or billions of data pairs) and obtain strong zero-shot capability. They adopt a simple yet effective contrastive objective that attracts paired image and caption and repels unpaired ones. There have been several additional works following the line of CLIP/ALIGN~\cite{zhai2021lit}. Florence~\cite{yuan2021florence} and BASIC \cite{pham2021combined} scale up the dataset and training with various backbones. FILIP~\cite{yao2021filip} focuses on generalizing the contrastive loss to local tokens for fine-grained supervision. DeCLIP~\cite{li2021supervision}, SLIP~\cite{mu2021slip} and other recent works extend supervision signal from self-supervision, multi-view supervision, nearest-neighbor supervision, object detections~\cite{zhong2022regionclip}, or external language knowledge~\cite{li2022clip}.   
Orthogonal to above mentioned works, this work focuses on the sharing of weights across vision and text modalities in large-scale contrastive pre-training.

\paragraph{Vision and Language Modeling:}
Another similar line of work is Vision-and-Language Pre-training (or VLP)~\cite{lu2019vilbert,tan2019lxmert,zhou2020unified,chen2019uniter,li2019visualbert,li2020unimo,wang2021vlmo,wang2021simvlm,li2022blip,wang2021ufo,wang2022multimodal}, where both vision and language signals are fed into also a unified model to enable downstream multi-modal tasks. Moreover, \cite{nagrani2021attention} utilizes a set of shared tokens across different modalities to enable multi-modal fusion. But there are two main differences between VLPs and this work: First, in VLP approaches, the model input consists of both image and text modalities concurrently, where the model attends to both modalities at the same time (essentially conducting modality fusion). In CLIP and MS-CLIP, the Transformer's input is either image or text individually: each modality is processed in isolation, where the two modalities are never processed concurrently (except for computing the contrastive loss at the end). Secondly, VLP works focus on designing unified fusion modules to blend multi-modal input well and target at multi-modal tasks (\textit{e.g.}, VQA, grounding), while the goal of our work is to allow parameter and knowledge sharing for uni-modal input and mainly serves visual-only downstream tasks.



\paragraph{Parameter-sharing Across Modalities:}
As humans reason over various modalities simultaneously, sharing modules for multi-modal processing has attracted increasing interests recently from the research community. \cite{lee2020parameter} proposes to share the parameters of Transformers across both layers and modalities to save parameters. They focus on video-audio multi-modal downstream tasks and have an additional multi-modal Transformer for modality fusion. 
\cite{lu202012} proposes to train a fully shared Multi-modal Transformer on 12 vision-language datasets. \cite{hu2021unit} further introduces a shared Transformer decoder for multi-task multi-modal learning. The most relevant work to ours is VATT \cite{akbari2021vatt}. VATT introduces a modality-agnostic transformer that can process video, text, and audio input and is pre-trained on a contrastive objective. The proposed model naively reuses the entire network for all modalities and yields results worse than the non-shared counterpart.
In contrast, this work studies more than whether a model can be shared, but rather how various degrees of sharing and design nuances behave, and which of those design choices might be useful to improve performance. 

\section{Methods}
\label{sec:methods}



\begin{figure*}[t]
\centering
\includegraphics[width=1\linewidth]{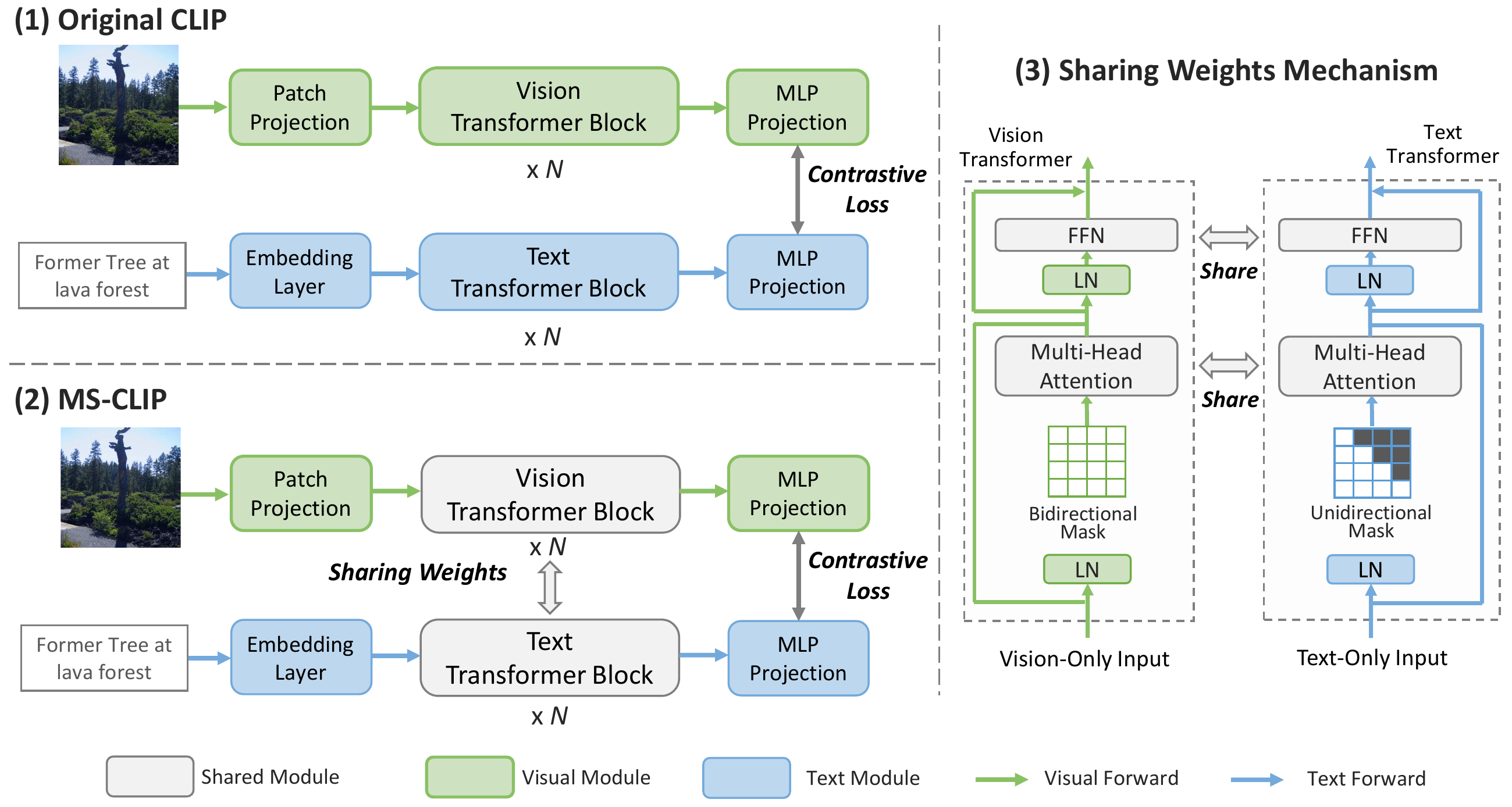}
\caption{Overview of the (1) vanilla CLIP and (2) our proposed baseline MS-CLIP, and (3) details of sharing mechanism MS-CLIP. }
\label{fig:diagram}
\end{figure*}

\subsection {Sharable Modules}
\label{sec:shareable_modules}
Following \cite{radford2021learning},  we use the Vision Transformer as the basic vision encoder (ViT-B/32 by default), and the transformer encoder as the basic text encoder, as shown in Fig. \ref{fig:diagram}-1. The challenge is to merge these two architectures. To accomplish this, we adjust the hidden dimension of text transformer from 512 to 768 to match that in the vision transformer. The resulting additional baseline method is noted as \CLIPb. After the adjustment, the resulting shared encoder uses 12 layers, with the vast majority of parameters able to be shared between two modalities, including the \textbf{attention modules}, \textbf{feedforward modules}, and \textbf{LayerNorm (LN) layers}. Modules that cannot be shared include the input embedding layer (where the vision encoder deploys a projection layer to embed image patches, while the text encoder encodes word tokens), and the output projection layer.  

We performed an experimental analysis to examine the impact of various degrees of weight sharing across modalities (see Sec. \ref{sec:sharing}: {\em On Modality-Shared Components}). In summary, the observations of that study are as follows: (1) LNs need to be modality-specific while the rest can be modality-shared; (2) Sharing all layers is better than a subset. Subsequently, a model sharing the attention and feedforward modules, while keeping the LNs modality specific, across all 12 layers, is regarded as the baseline of our model family.   We dub this Na\"\i ve modality sharing model MS-CLIP (see Fig. ~\ref{fig:diagram}-2 and  ~\ref{fig:diagram}-3).


\subsection{Modality-Specific Auxiliary Module Architecture}
\label{sec:modality_specific}

\begin{figure*}[t!]
\centering
\includegraphics[width=1.0\linewidth]{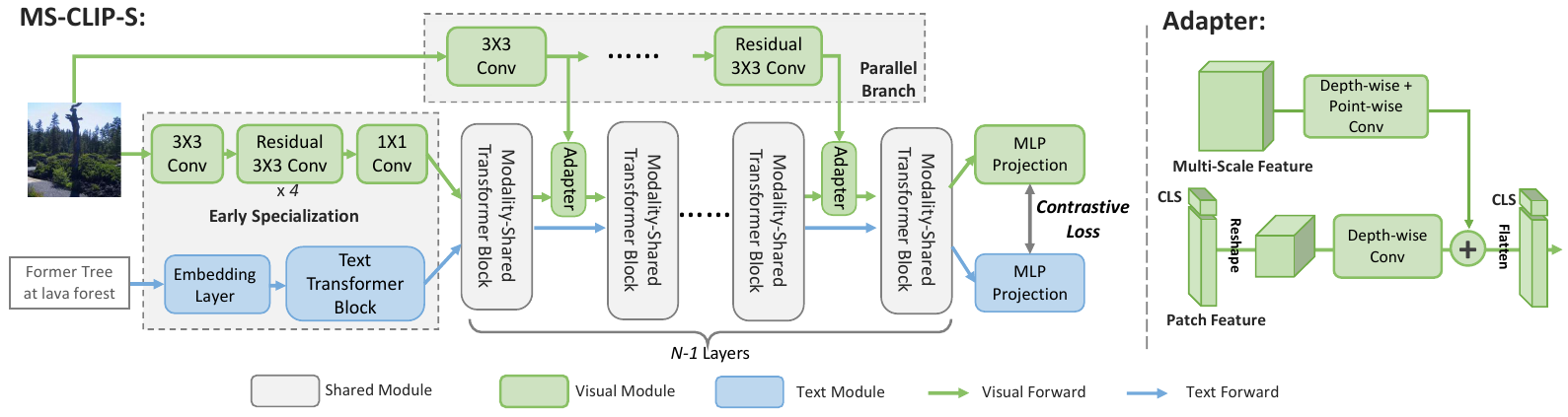}
\caption{Overview of MS-CLIP-S. Based on MS-CLIP, Early Specialization is further introduced at the beginning of the network. Simultaneously, an Efficient Parallel Branch is integrated to provide modality-specific multi-scale features to main modality-shared branch. See Section \ref{sec:modality_specific} text for more details. }
\label{fig:diagram2}
\end{figure*}

In this section we describe modifications introducing two lightweight modality-specific auxiliary modules, shown in Fig. \ref{fig:diagram2}. We name the full model with both modality-specific designs as MS-CLIP-S, where ``S'' indicates ``Specialized branches''.

\paragraph{Early Specialization:}
\label{sec:early_spec}
The first modality-specific design specializes only the first layer for visual and text modalities, leaving other layers shared. Concretely, on vision side, we employ a series of convolutional networks with residual connections as our specialization layer, in which the feature resolution is down-sampled and the channel dimension is increased. The detailed configuration is shown in Tab. \ref{tab:earlyconv_config}, with ViT-B/32 as the visual encoder , inspired by~\cite{xiao2021early}. For other visual encoders, such as ViT-B/16, the configuration only differs in the strides of convolutions (see Supplement). 
We further add residual connections between convolutional layers, which is empirically more stable for large-scale training. On the language side, we reuse the de-facto Transformer layer for language modeling. 

\begin{figure*}[t]
\begin{minipage}[ht!]{\textwidth}
 \begin{minipage}[ht!]{\dimexpr.4\textwidth-1em}
  \centering
     \makeatletter\def\@captype{table}\makeatother\caption{Setting of Early Specialization. $N{\times} N$ signifies the 2D kernel size of CNNs.}
     \label{tab:earlyconv_config}
        \fontsize{7}{7}\selectfont
       \begin{tabular}{cccc}
        \toprule
        Module& Dim& Resolution\\
        \midrule
        $3{\times}3$ Conv&3$\rightarrow$48&224$\rightarrow$112\\
        Residual $3{\times}3$ Conv&48$\rightarrow$96&112$\rightarrow$56\\
        Residual $3{\times}3$ Conv&96$\rightarrow$192&56$\rightarrow$28\\
        Residual $3{\times}3$ Conv&192$\rightarrow$384&28$\rightarrow$14\\
        Residual $3{\times}3$ Conv&384$\rightarrow$768&14$\rightarrow$7\\
        $1{\times}1$ Conv&768$\rightarrow$768&7$\rightarrow$7\\
        \midrule
        Total \# Parameters&\multicolumn{2}{c}{4.5M}\\
        \bottomrule
    \end{tabular}
  \end{minipage}
    \hfill
  \begin{minipage}[ht!]{\dimexpr.6\textwidth-1em}
   \centering
        \makeatletter\def\@captype{table}\makeatother\caption{Setting of Efficient Parallel Branch. Fusion Layer means fusing with which modality-shared layer.}
        \label{tab:parallel_config}
        \fontsize{7}{7}\selectfont
         \begin{tabular}{ccccl}
            \toprule
            Parallel & Adapter & Fusion & Resol-\\
            Module & Module & Layer &ution\\
            \midrule
            $3{\times}3$ Conv&$16{\times}16$ DWConv&2&224$\rightarrow$112\\
            Bottleneck $3{\times}3$ Conv&$8{\times}8$ DWConv&4&112$\rightarrow$56\\
            Bottleneck $3{\times}3$ Conv&$4{\times}4$ DWConv&6&56$\rightarrow$28\\
            Bottleneck $3{\times}3$ Conv&$2{\times}2$ DWConv&8&28$\rightarrow$14\\
            Bottleneck $3{\times}3$ Conv&$1{\times}1$ DWConv&10&14$\rightarrow$7\\
            \midrule
            Total \# Parameters&\multicolumn{3}{c}{3.9M}\\
            \bottomrule
      \end{tabular}
   \end{minipage}
\end{minipage}
\end{figure*}

\paragraph{Efficient Parallel Branch:}
For image representations, multi-scale information has been demonstrated to be valuable ~\cite{cai2016unified,szegedy2015going}. Vision Transformers \cite{dosovitskiy2020image}, however, typically operate on a fixed scale. In recent works that introduce multi-scale information into ViT \cite{liu2021swin,wu2021cvt}, the patch size is gradually reduced and the dimension of the channel is increased, stage by stage. Nevertheless, directly sharing weights between multi-scale ViT and the language Transformer is non-trivial, due to the discrepancy in their channel dimensions. Motivated by \cite{feichtenhofer2019slowfast}, we propose to have an auxiliary parallel vision branch alongside the shared Transformer, which consists of one convolution layer and four residual convolution layers, to decrease the resolution and increase the channel dimension (see Fig. \ref{fig:diagram2}). In contrast with the plain residual convolutions in Early Specialization, here we utilize the bottleneck design in ResNet \cite{he2016deep} to be parameter efficient. The main function of the parallel branch is to supplement the shared branch with multi-scale features for image information. Therefore, we employ one adapter after each parallel layer to integrate features from varying scales into layers of the shared Transformer. For further efficiency, we adopt depth-wise convolutions (DWConv) and point-wise convolution (PWConv) in adapters to adjust the feature map size and depth.  The adapter can be formulated as:

\begin{equation}
\begin{aligned}
    H^{'}_{p} = bn(\text{PWConv}(\text{DWConv}(H_{p}))) \\
    H^{'} = ln(bn(\text{DWConv}(H)) + H^{'}_{p}),
    \label{eq:adapter}
\end{aligned}
\end{equation}

\noindent where $H_{p}$ is the multi-scale feature in parallel branch, and $(H, H^{'})$ is the adapter's input and output, respectively. $bn$ and $ln$ denote batch normalization and layer normalization. Note that the \textit{CLS} token is not fused with parallel branch and remains unchanged. The outputs of 5 parallel layers are fused with every other shared Transformer layers.  The detailed configuration when ViT-B/32 being visual encoder is provided in Tab. \ref{tab:parallel_config}. For other visual encoders, such as ViT-B/16, only the kernel size and stride differs, and we attach the configuration in Supplementary.

\section{Experiments}
\label{sec:adapters}

Section \ref{sec:setup} introduces the pre-training and evaluation setup. Sections \ref{sec:expres} and \ref{sec:ablation} details the primary experimental results and related ablations. Section \ref{sec:laion} presents experiments where the pretraining data is changed. Finally, Section \ref{sec:furtheranalysis} presents experiments to better elucidate why MS-CLIP works.

\subsection{Setup}
\label{sec:setup}
\paragraph{Training Details:} Similar to the original CLIP paper  \cite{radford2021learning}, we maintain separate attention masks for image and text: vision transformer allows upper layers to attend to all tokens from lower layers with a bi-directional mask, while the mask in text transformer is auto-regressive. The optimizer is AdamW \cite{loshchilov2017decoupled}. The learning rate is decayed from 1.6e-3 to 1.6e-4, with a cosine scheduler and a warm up at first 5 epochs. We train our models on 16 NVIDIA V100 GPUs with the batch size per GPU set to be 256. 
For MS-CLIP and MS-CLIP-S, the weight decay for non-shared parameters and shared parameters are separately set to 0.05 and 0.2. We found that a higher weight decay for shared parameters works better, simply because shared parameters are updated twice in each iteration, and a higher weight decay can mitigate over-fitting.

\paragraph{Pre-training Dataset:} By default, we use YFCC100M \cite{thomee2016yfcc100m} for pre-training. Following the filtering process in \cite{radford2021learning}, we only keep image-text pairs where captions are in English. This leaves us around 22 million data pairs. 
All our results are reported on this data version, including the vanilla CLIP ~\cite{radford2021learning}. Subsequently, we also pre-train both our model and vanilla CLIP on a subset of the more recent dataset: LAION-400M \cite{schuhmann2021laion}. More details can be found in Sec. \ref{sec:laion}.

\paragraph{Evaluation Datasets:} 
In total, we adopt 25 public datasets for evaluation by either zero-shot learning or linear probing: ImageNet~\cite{deng2009imagenet}, Food-101 \cite{bossard14}, CIFAR-10 \cite{krizhevsky2009learning}, CIFAR-100 \cite{krizhevsky2009learning}, SUN397 \cite{xiao2010sun}, Stanford Cars \cite{krause20133d}, FGVC Aircraft \cite{maji2013fine}, Pascal Voc 2007 Classification \cite{pascal-voc-2007}, Describable Texture (DTD) \cite{cimpoi14describing}, Oxford-IIIT Pets \cite{parkhi2012cats}, Caltech-101 \cite{fei2004learning}, Oxford Flowers 102 \cite{nilsback2008automated}, MNIST \cite{lecun1998gradient}, Facial Emotion Recognition (FER) \cite{pantic2005web}, STL-10 \cite{coates2011analysis}, GTSRB \cite{stallkamp2012man}, PatchCamelyon \cite{Veeling2018-qh}, UCF101 \cite{soomro2012ucf101}, Hateful Memes \cite{kiela2020hateful}, Country211 \cite{radford2021learning}, EuroSAT \cite{helber2019eurosat}, Kitti-distance \cite{geiger2012we}, Rendered-SST2 \cite{socher2013recursive}, Resisc45 \cite{cheng2017remote}, MSCOCO \cite{lin2014microsoft}. These datasets cover various categories, including generic objects, memes, scenes and etc. We perform linear probing with logistic regression on top of extracted image features, exactly following the protocol in the original CLIP paper~\cite{radford2021learning}. For zero-shot recognition, we report zero-shot accuracy on the ImageNet \cite{deng2009imagenet} validation set. Following CLIP, we use an ensemble of multiple prompts to extract text features as category features. For zero-shot image-text retrieval, we report recall on MSCOCO \cite{lin2014microsoft}

\subsection{Experimental Results}
\label{sec:expres}
\paragraph{Compared Models:} We conduct experiments on proposed MS-CLIP-S and vanilla CLIP \cite{radford2021learning}. Both ViT-B/32 and ViT-B/16 are adopted as visual encoders. As stated in Sec \ref{sec:setup}, we strictly follow the implementation in \cite{radford2021learning}. 


\begin{table*}[t!]
\caption{Experimental results of zero-shot image classification (ZS*), linear probing and zero-shot image-text retrieval (ITR*) across 25 datasets.}
\label{tab:main_result}
\fontsize{8.5}{8.5}\selectfont
\begin{center}
\begin{tabular}{llllccc|ccc}
\toprule
\multirow{2}{*}{Eval.} &\multicolumn{3}{l}{\multirow{2}{*}{Datasets}}  & CLIP & MS-CLIP-S&\multirow{2}{*}{$\Delta$}& CLIP & MS-CLIP-S&\multirow{2}{*}{$\Delta$}\\
&  &  &  &(ViT-B/32) & (ViT-B/32) & &(ViT-B/16) & (ViT-B/16) &\\
\midrule
\multirow{24}{*}{\rotatebox[origin=c]{90}{Linear Probing}}&\multicolumn{3}{l}{Food-101}&71.3&\textbf{\underline{76.0}}&$+$4.7&80.1&81.5&$+$1.4\\
&\multicolumn{3}{l}{SUN397}&68.1&\textbf{\underline{71.7}}&$+$3.6 & 72.3 & \textbf{\underline{73.2}} &$+$0.9\\
&\multicolumn{3}{l}{Stanford Cars}&21.8&\textbf{\underline{27.5}}&$+$5.7 & 27.6 & \textbf{\underline{32.0}} &$+$4.4\\
&\multicolumn{3}{l}{FGVC Aircraft}&31.8&\textbf{\underline{32.9}}&$+$1.1 & 33.6 & \textbf{\underline{38.4}} &$+$4.8\\
&\multicolumn{3}{l}{Pascal Voc 2007}&84.4&\textbf{\underline{86.1}}& $+$1.7& 85.6 & \textbf{\underline{86.7}} &$+$1.1\\
&\multicolumn{3}{l}{DTD}&64.1&\textbf{\underline{69.4}}&$+$5.3 & 67.6 & \textbf{\underline{71.9}} &$+$4.3\\
&\multicolumn{3}{l}{Oxford-IIIT Pets}&61.1&\textbf{\underline{62.1}}&$+$1.0& 63.0 & \textbf{\underline{63.7}} &$+$0.7\\
&\multicolumn{3}{l}{Caltech-101}&\textbf{\underline{82.8}}&81.6&$-$1.2 & 83.6 & \textbf{\underline{83.8}} &$+$0.2\\
&\multicolumn{3}{l}{Oxford Flowers 102}&90.7&\textbf{\underline{93.8}}&$+$3.1 & 94.0 & \textbf{\underline{95.2}} &$+$1.2\\
&\multicolumn{3}{l}{MNIST}&96.5&\textbf{\underline{97.2}}&$+$0.7& \textbf{\underline{96.9}} & 96.7 &$-$0.2\\
&\multicolumn{3}{l}{FER}&\textbf{\underline{54.9}}&53.6&$-$1.3 & 55.3 & \textbf{\underline{56.2}} &$+$0.9\\
&\multicolumn{3}{l}{STL-10}&\textbf{\underline{95.4}}&95.1&$-$0.3 & \textbf{\underline{96.9}} & 96.7 &$-$0.2\\
&\multicolumn{3}{l}{GTSRB}&67.1&\textbf{\underline{69.9}}&$+$2.8 & 72.5 & \textbf{\underline{78.3}} &$+$5.8\\
&\multicolumn{3}{l}{PatchCamelyon}&78.3&\textbf{\underline{81.3}}&$+$3.0 & \textbf{\underline{82}} & 80.4 &$-$1.6\\
&\multicolumn{3}{l}{UCF101}&72.8&\textbf{\underline{74.6}}&$+$1.8 & 74.6 & \textbf{\underline{75.3}} &$+$0.7\\
&\multicolumn{3}{l}{CIFAR-10}&\textbf{\underline{91.0}}&87.2&$-$3.8 & \textbf{\underline{91.1}} & 89.8 &$-$1.3\\
&\multicolumn{3}{l}{CIFAR-100}&\textbf{\underline{71.9}}&66.7&$-$5.2 & \textbf{\underline{72.6}} & 71.5 &$-$1.1\\
&\multicolumn{3}{l}{Hateful Memes}&50.6&\textbf{\underline{52.4}}&$+$1.8 & \textbf{\underline{51.6}} & 50.2 &$-$1.4\\
&\multicolumn{3}{l}{ImageNet}&58.5&\textbf{\underline{63.7}}&$+$5.1 & 64.7 & 66.7 &$+$2.0\\
&\multicolumn{3}{l}{Country211}&19.9&\textbf{\underline{21.9}}&$+$2.0 &23.5 & \underline{23.6} &$+$0.1\\
&\multicolumn{3}{l}{EuroSAT}&\textbf{\underline{94.4}}&93.5&$-$0.9  & \textbf{\underline{94.6}} & 94.3 &$-$0.3\\
&\multicolumn{3}{l}{Kitti-distance}&39.7&\textbf{\underline{45.1}}&$+$5.4 & 35.7 & \textbf{\underline{40.2}} &$+$4.5\\
&\multicolumn{3}{l}{Rendered-SST2}&55.2&\textbf{\underline{56.0}}&$+$0.8 & 56.8 & \textbf{\underline{56.9}} &$+$0.1\\
&\multicolumn{3}{l}{Resisc45}&83.3&\textbf{\underline{85.1}}&$+$1.8 & 85.6 & \textbf{\underline{86.5}} &$+$0.9\\
\cmidrule{2-10}
&\multicolumn{3}{l}{Avg.}&66.9&\textbf{\underline{68.5}}&$+$1.6 & 69.2 & \textbf{\underline{70.4}} &$+$1.2\\
\midrule
ZS* &\multicolumn{3}{l}{ImageNet} &32.2&\textbf{\underline{36.7}}&$+$4.5 & 36.9 & \textbf{\underline{39.0}} &$+$2.1\\
\midrule
\multirow{4}{*}{ITR*} & \multirow{4}{*}{MSCOCO} &\multirow{2}{*}{I2T} & R@1 &24.4&\textbf{\underline{28.5}}&$+$4.1 &27.5&\textbf{\underline{29.9}}&$+$2.4\\
 & &&R@5&48.5&\textbf{\underline{54.1}}&$+$5.6 &51.9&\textbf{\underline{56.8}}&$+$4.9\\
 & & \multirow{2}{*}{T2I}&R@1&14.8&\textbf{\underline{19.4}}&$+$4.6&17.7&\textbf{\underline{20.4}}&$+$2.7\\
 & &&R@5&34.9&\textbf{\underline{40.8}}&$+$5.9 &38.7&\textbf{\underline{42.9}}&$+$4.2\\

\bottomrule
\end{tabular}
\end{center}
\end{table*}



\paragraph{Zero-Shot ImageNet Classification:} The experimental results are reported in the row of ZS* in Tab. \ref{tab:main_result}. 
By comparing the four columns, we find that, based on ViT-B/32 (ViT-B/16), MS-CLIP-S can outperform CLIP by 4.5 (2.1) percentage points, or 13.9\% (5.6\%) relative, in zero-shot recognition accuracy on ImageNet. 


\paragraph{Linear Probing:} To fully compare our model with vanilla CLIP, the results of linear probing on 24 various datasets are shown in Tab.~\ref{tab:main_result}. Overall, with ViT-B/32 (ViT-B/16) as backbone, MS-CLIP-S outperforms vanilla CLIP on 18 (17) out of 24 tasks, and the average improvement on 24 tasks is 1.62 (1.16) points.

\paragraph{Zero-shot Image-Text Retrieval:} We evaluate our MS-CLIP-S on two sub-tasks: image-to-text retrieval and text-to-image retrieval under zero-shot setting. The dataset we used is MSCOCO test set, which has 5,000 images. The comparison between MS-CLIP-S and vanilla CLIP, both pre-trained on YFCC, is shown in the last 4 rows of Tab. \ref{tab:main_result}. With both ViT-B/32 and ViT-B/16, our MS-CLIP-S outperforms vanilla CLIP by a large margin across the board.

\subsection{Ablation Study} 
\label{sec:ablation}
For the following ablation analysis, we use ViT-B/32, and report zero-shot accuracy on ImageNet validation set.
\subsubsection{On Modality-Shared Components:}  
\label{sec:sharing}
We systematically study the impact  of  varying  the  degree  of  sharing  of  components  across  different  layers, and make the following observations:

\begin{table*}[t]
\caption{Experimental results of sharing different components in Transformer layer. First two rows are baselines without sharing. LN1 denotes the LN before Attn. LN2 denotes the LN before FFN.}
\label{tab:share_ln}
\begin{center}
\begin{tabular}{ccccc}
\toprule
Text&\multirow{2}{*}{\# Params}&Shared &Non-Shared &Zero-shot \\
Width& &Module& Module&Acc(\%) \\
\midrule
512&151M&-   & Attn, FFN, LN1, LN2 & 32.15\\
768&209M&-    & Attn, FFN, LN1, LN2 & 31.85\\
768&126M &Attn, FFN, LN1, LN2   & - & 28.40\\
768&126M &Attn, FFN, LN1    & LN2 & 27.57\\
768&126M &Attn, FFN, LN2    & LN1 & 32.16\\
768&126M &Attn, FFN    & LN1, LN2 & \textbf{32.99}\\
\bottomrule
\end{tabular}
\end{center}
\end{table*}

1. \textbf{LNs need to be modality-specific. }
We examine the shareable modules within each Transformer layer, excluding the input and output projection layers, which cannot be shared. As shown in Tab. \ref{tab:share_ln}, the first model variant shares all components (across all layers for simplicity), including two LN layers and transformation weights in the self-attention and feedforward modules, which results in worse performance (28.4\%) compared to \CLIPa{} (32.15\%) and \CLIPb (31.85\%). Then we examine making the two LN layers modality-specific, which yields better performance in both zero-shot accuracy (32.99\%) and parameter efficiency. Note that the number of parameters in LNs is negligible compared with the transformation weights. Our observation echos the finding in FPT~\cite{lu2021pretrained} that only tuning LNs in a mostly-frozen pre-trained language model yields satisfactory performance on vision tasks.

\begin{table*}[t]
\caption{Results of sharing different layers in Transformer.}
\label{tab:share_layer}
\fontsize{9.5}{9}\selectfont
\begin{center}
\begin{tabular}{ccccccccccc}
\toprule
Share Last X layers&12&11&10&8&6&4&2&0&NAS-Search\\
\midrule
Zero-shot Acc(\%)&32.99&31.25&32.21&32.39&32.85&30.91&nan&31.85&30.97\\
\# Parameters &126M&132M&139M&153M&167M&181M&195M&209M&174M \\ 
\bottomrule
\end{tabular}
\end{center}
\end{table*}

2. \textbf{Less is more: Sharing all layers is better than some.}
We further study which layer should be modality-specific and which should be modality-shared. We conduct experiments on sharing the last N layers where $N$ is ranging from 12 to 0. $N=12$ indicates all layers are shared and $N=0$ indicates the non-shared baseline \CLIPb. Tab. \ref{tab:share_layer} suggests that sharing all 12 layers performs the best while requiring the least number of parameters. This design of sharing all layers is what we refer to as MS-CLIP. Additionally, inspired by recent work on Neural Architecture Search (NAS) \cite{zheng2021migo,dong2019searching}, we train a model that learns a policy to control which layer to (not) share via Gumbel Softmax \cite{dong2019searching}.
Despite its sophistication, it still underperforms MS-CLIP.

\subsubsection{On Modality-Specific Designs:}  
We conduct experiments with the following settings: (1) \CLIPa{}: The same as \cite{radford2021learning}, this uses ViT-B32 as the visual encoder, and Text Transformer with width set to 512. (2) \CLIPb{}: This model sets the width of Text Transformer as 768 to unify the dimension of both encoders. (3) \MACLIPa{}: Compared with \CLIPb{}, this model utilizes the modality-shared transformer blocks to substitute non-shared transformer blocks in visual and text encoders. We use the best setting found in Sec. \ref{sec:sharing}: sharing all except for two layer normalizations. (4) \MACLIPb{}: Based on (3), we specialize the first layer of shared visual \& text encoders following Sec. \ref{sec:methods}. (5) \MACLIPa{} + Parallel Branch: Based on (3), we add a parallel branch to shared visual encoder. (6)  \MACLIPd{}: Based on (3), we apply both early specialization and parallel branch to our shared visual \& text encoders. 

The result is summarized in Tab.~\ref{tab:zeroshot}.  By comparing the 2nd row and the 3rd row, we find that directly increasing the capacity of the text transformer yields worse results.
Then comparing 3-rd row and 4-th row, we find that sharing parameters in vision and text transformer improves the performance and even can outperform \CLIPa{} (as also shown in previous ablation on modality-shared modules). 
Comparing 4th and 5th row against the 1st row, we notice that
early specialization can contribute to 2.1\% improvement with only a 4M parameters increase and auxiliary parallel branch on vision has a 1.1\% boost. The full model in 6th row further advances to 36.66\%, a 4.5\% absolute gain over the baseline \CLIPa{}.



\begin{table}[t]
\caption{Ablation on Modality-Sepcific Designs.}
\label{tab:zeroshot}
\begin{center}
\begin{tabular}{lcc}
\toprule
Module&\multirow{2}{*}{\# Parameters}&Zero-shot \\
Name& &Acc(\%) \\
\midrule
\CLIPa{}&151M& 32.15\\
\CLIPb{}&209M& 31.85\\
\MACLIPa{}&126M& 32.99\\
  $\cdots$   w/ Early Specialization&129M& 35.18\\
  $\cdots$   w/ Parallel Branch&129M& 34.18\\
\MACLIPd{} &132M& \textbf{36.66}\\
\bottomrule
\end{tabular}
\end{center}
\end{table}

\begin{table}[t]
\caption{Results of models pre-trained on Laion-20M: zero-shot image classification, linear probing and zero-shot image-text retrieval (ITR*).}
\label{tab:laion}
\begin{center}
\begin{tabular}{c|c|c|c|c|c|c|c}
\toprule
&ImageNet&\multicolumn{4}{c|}{MSCOCO Test ITR*}&\multicolumn{2}{c}{Linear Probing}\\
&Zero-shot&\multicolumn{2}{c|}{I2T}&\multicolumn{2}{c|}{T2I}&\multicolumn{2}{c}{on 24 datasets}\\
&Acc(\%)&R@1 &R@5 &R@1 &R@5& Average & \#Wins \\
\midrule
CLIP&35.5&24.7&48.1&16.2&35.8&70.5 & 5\\
\midrule
MS-CLIP-S&\textbf{\underline{40.2}}&\textbf{\underline{31.2}}&\textbf{\underline{57.4}}&\textbf{\underline{20.6}}&\textbf{\underline{43.6}}&\textbf{\underline{73.3}} & \textbf{\underline{19}}\\
\midrule
$\Delta$&$+$4.7&$+$6.5&$+$9.3&$+$4.4&$+$7.8&$+$2.8 &$+$14\\
\bottomrule
\end{tabular}
\end{center}
\end{table}

\subsection{Pre-training Data Quality}
\label{sec:laion}
To verify that our proposed model can generalize to pre-training datasets of various quality, we pre-train both vanilla CLIP (ViT-B/32) and MS-CLIP-S (ViT-B/32) on a subset of the recently released public Laion-400M dataset~\cite{schuhmann2021laion}. This proof-of-concept subset contains 20M randomly-sampled image-caption pairs from Laion-400M, similar to the size of filtered YFCC. We name it Laion20M. The complete experimental results are shown in Tab. \ref{tab:laion}, where our model outperforms vanilla CLIP substantially. Since in the building of Laion-400M, a pre-trained CLIP is used to filter out the noisy image-text pairs, the dataset is believed to have higher data quality. This can also be proved by comparing vanilla CLIP's results in Laion and YFCC. Comparing Tab. \ref{tab:laion} and Tab. \ref{tab:main_result} side by side, we find that the improvement brought by proposed MS-CLIP-S  pre-trained on Laion-20M is generally higher than on YFCC (22M). It might imply that our method can benefit more when pre-training data quality is higher. Detailed performance of linear probing on 24 datasets is added in Supplementary.

\subsection{Further Analysis}
\label{sec:furtheranalysis}
There are likely multiple reasons that explain observed improvements in performance. Firstly, sharing the majority of parameters across vision and language can implicitly encourage the model to focus on the common pattern across two modalities and alleviate overfitting of trivial vision (\textit{e.g.}, illumination) or language cues (\textit{e.g.} stop words). Additionally, the auxiliary modality-specific modules, {\em Early Specialization} and {\em Parallel Branch},  provide vision-specific multi-scale features and language-specific features to complement the shared modules. To have an in-depth understanding, we perform the following further analysis:

\begin{table*}[t]
\caption{Layer-wise NMI scores of models. }
\fontsize{5.5}{5.5}\selectfont
\setlength{\tabcolsep}{2.1pt}
\label{tab:nmi_1}
\begin{center}
\begin{tabular}{lcccccccccccc|c}
\toprule
Layer &0&1&2&3&4&5&6&7&8&9&10&11&Avg.\\
\midrule
\CLIPb{} &0.586&0.387&0.265&0.252&0.255&0.241&0.239&0.243&0.235&0.23&0.227&0.185&0.278\\
\MACLIPa{} &0.589&0.332&0.235&0.211&0.2&0.21&0.2&0.202&0.214&0.197&0.192&0.173&0.246 \\ 
 $\cdots$   w/ Early Specialization &0.471&0.348&0.215&0.21&0.218&0.221&0.22&0.213&0.19&0.183&0.179&0.161&\textbf{0.235} \\ 
\MACLIPd{} &0.519&0.536&0.243&0.216&0.199&0.221&0.19&0.247&0.216&0.215&0.224&0.217&0.270 \\ 
\bottomrule
\end{tabular}
\end{center}
\setlength{\tabcolsep}{6pt}
\end{table*}

\paragraph{NMI Score: Shared model exhibits higher multi-modal fusion degree.} To probe the multi-modal fusion degree, following \cite{cao2020behind}, we measure the Normalized Mutual Information (NMI) between visual features and text features at each layer.
For each image-caption pair, we use K-means algorithm (K=2) to group all feature vectors from the forward pass of visual input and text input into 2 clusters. Then, NMI is applied to measure the difference between the generated clusters and ground-truth clusters. The higher the NMI score is, the easier the visual features and text features can be separated, and the lower the multi-modal fusion degree is. 

NMI scores are then used to probe the multi-modal fusion degree of the shared model (\MACLIPa{}) vs. non-shared model (\CLIPb{}).  Here we choose \CLIPb{} instead of \CLIPa{} in that the feature dimensions of two modalities have to be the same for clustering. The measurement is performed on randomly sampled 50k image-caption pairs from YFCC100M dataset. NMI scores of all 12 layers and the average are listed in the first two rows of Tab.~\ref{tab:nmi_1}. Shared model has lower NMI scores than original CLIP on almost all the layers and the average, indicating a higher degree of multi-modal fusion. 

Following the same procedure as above, we further report the NMI scores of \MACLIPb{} and \MACLIPd{} (see Tab.~\ref{tab:nmi_1}). The result shows that sharing parameters and introducing early specialization can improve the multi-modal fusion degree, which coincides with our hypothesis mentioned above. However, adding parallel branch leads to a lower fusion score. 
This is somewhat conflicting with what we see in Tab. \ref{tab:zeroshot}, where adding parallel branch enhances the learned representation. 
In the following subsection, we explore other metrics to further probe into what contributes to this behavior.  



\begin{table*}[t]
\caption{ Common Semantic Structure distance}
\fontsize{5.5}{5.5}\selectfont
\setlength{\tabcolsep}{2.1pt}
\label{tab:semantic_similarity}
\begin{center}
\begin{tabular}{lcccccccccccc|c}
\toprule
Layer &0&1&2&3&4&5&6&7&8&9&10&11&Avg.\\
\midrule    
\CLIPa{}&0.18&0.203&0.227&0.186& 0.178&0.164&0.118&0.103&0.106&0.109&0.105&0.074&0.143\\
\MACLIPa{} &0.175&0.128&0.153&0.132&0.136&0.136&0.106&0.119&0.092&0.106&0.083&0.058&0.113 \\ 
 $\cdots$ w/ Early Specialization &-&0.107&0.142&0.16&0.12&0.12&0.103&0.103&0.096&0.111&0.11&0.058&0.111 \\ 
\MACLIPd{} &-&0.085&0.162&0.105&0.102&0.103&0.105&0.114&0.093&0.094&0.093&0.061&\textbf{0.101} \\ 

\bottomrule
\end{tabular}
\end{center}
\setlength{\tabcolsep}{6pt}
\end{table*}

\paragraph{Multi-modal Common Semantic Structure: The Integration of Modality-Shared and Modality-Specific modules learns better common patterns.} 
One of the hypotheses on why MS-CLIP architectures perform better is that they better capture the common semantic structures inherent to concepts from different modalities.

\begin{wrapfigure}{l}{0.42\textwidth}
        \centering
        \includegraphics[width=.28\textheight]{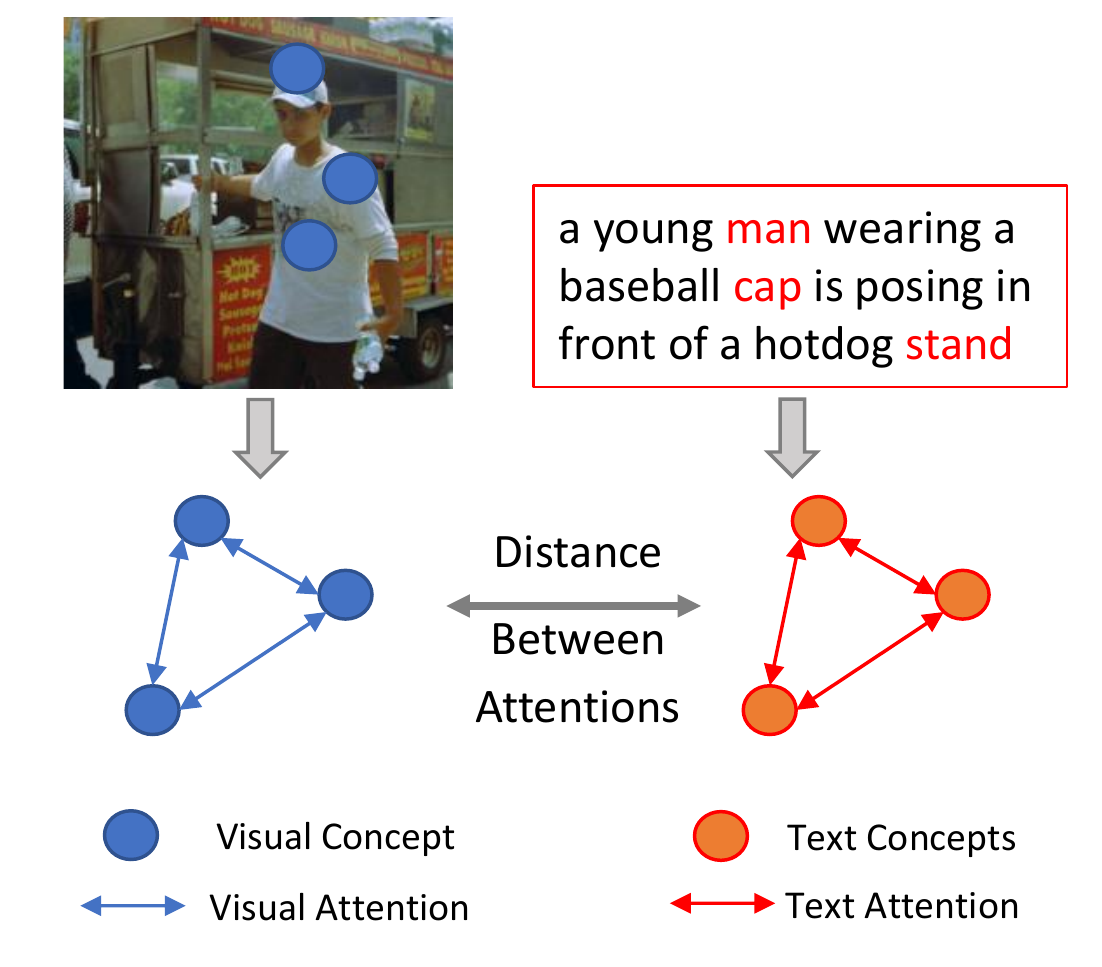}
        \caption{Diagram of computing Common Semantic Structure distance.}
\label{fig:common_semantic_structure}
\end{wrapfigure}

To justify this hypothesis, we propose to measure the similarity between attention weights on visual concepts and the corresponding language concepts (see Fig.~\ref{fig:common_semantic_structure}). The measurement is performed on a surrogate dataset named Flick30K-Entities~\cite{plummer2015flickr30k}, where object regions in each image are grounded to their corresponding phrases in a caption, in the form of bounding boxes. Given an image, assuming there are grounded object regions (visual concepts) $\{vc_1, vc_2, ..., vc_n\}$ and corresponding object words (language concepts)  $\{tc_1, tc_2, ..., tc_n\}$, where $tc_i$ is associated with $vc_i$ semantically.  In the $h$-th head of $l$-th attention layer, we denote the raw visual attention matrix as $M^{lh}$ and the raw text attention matrix as $K^{lh}$. 
We then regard the attention value between $tc_i$ and $tc_j$ as $K^{lh}_{ij}$, and attention value between $vc_i$ and $vc_j$ as $M^{lh}_{ij}$. 
We extract the attention values from concept $i$ to all other concepts (i.e., $j$) and normalize for visual attention and language attention, respectively (denoted as ``attention vectors'').
The final attention vectors are averaged over all heads in that attention layer. We compute the attention vectors for all concept pairs $i$. Finally, we measure the $l1$ distance between the visual attention vector and the language attention vector and sum them up over all the concept pairs and treat it as the Common Semantic Structure (CSC) distance of that attention layer. A lower CSC distance means more common attention patterns learned across modalities.  The whole process can be formulated as:

\begin{equation}
\begin{aligned}
    dis^{l}_{ij} = |\sum\limits_{h=1}^H \frac{1}{H} softmax_{i}(M^{lh}_{ij})-\sum\limits_{h=1}^H \frac{1}{H} softmax_{i}(K^{lh}_{ij})|
\end{aligned}
\end{equation}
\begin{equation}
\begin{aligned}
    CSC^{l} = dis^{l} = \sum\limits_{i=1}^n \sum\limits_{j=1}^n (dis^{l}_{ij}).
\end{aligned}
\end{equation}

The layer-wise  CSC distance of \CLIPa{}, \MACLIPa{}, \MACLIPb{} and \MACLIPd{} are reported in Tab. \ref{tab:semantic_similarity}. 
The first layers of \MACLIPb{} and  \MACLIPd{} are omitted as their vision branch do not contain any attention weights. The average score is computed based on the CSC distance on their last 11 layers. We find that 
our proposed modules lower the CSC distance and learn more modality-agnostic representation. Unsurprisingly, sharing parameters can enforce the attention to learn more common information. At the same time, it might reduce the overfitting brought by training separately. As for our proposed modality-specific modules, we suspect that these well designed modules can account for the discrepancy of individual modalities, especially by the vision-specific multi-scale feature, and thus facilitate the learning of the common patterns with the share component. 

\begin{figure*}[t]
\centering
\includegraphics[width=1.0\linewidth]{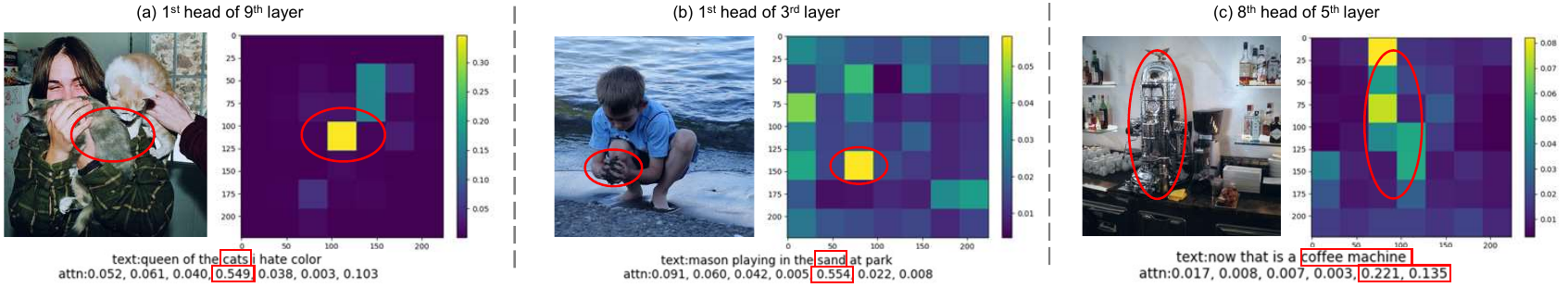}
\caption{Visualized attention maps of shared attention head (Please zoom in to see the caption). }
\label{fig:vis}
\end{figure*}

\paragraph{Visualization of Shared Attention Head:} In order to intuitively understand how shared attention module works, we visualize the visual attention patterns and text attention patterns of the same shared attention head during inference. More precisely, for vision, we visualize the attention weights at the final layer from the \textit{CLS} token to all its input tokens. For text, we perform the same except on the \textit{EOS} token. 
Note that both \textit{CLS} token and \textit{EOS} token are treated as the feature representation.
Results on \MACLIPd{} are shown in Fig.~\ref{fig:vis}. Interestingly, some heads are able to attend on the same concepts from different modalities. We  take Fig. \ref{fig:vis}(a) as an example. Given the image and caption respectively as input, the 1st head of 9th attention layer gives the highest attention value to the region of ``cat'' in image and token ``cats'' in text. It suggests the learning of co-reference across modalities.


\section{Conclusion}
\label{sec:conclusion}
We propose MS-CLIP, a modality-shared contrastive language-image pre-training approach, where most parameters in vision and text encoders are shared. To explore how many parameters/layers can be shared across modalities, we carefully investigate various architectural design choices through extensive experiments. In addition, we propose two modality-specific auxiliary designs: Early Specialization and Auxiliary Parallel Branch. Experiments on both zero-shot recognition and linear probing demonstrate the superiority of MS-CLIP architectures over the vanilla CLIP in both effectiveness and parameter efficiency. Finally, further analysis into the proposed architecture shows that sharing parameters can help map the two modalities into a closer embedding space and promote learning a common semantic structure.



\noindent{\bf Acknowledgement:}
This work is done during Haoxuan's internship at Microsoft. This work is also supported in part by DARPA MCS program under Cooperative Agreement N66001-19-2-4032.

\clearpage
%
%
\bibliographystyle{splncs04}
\bibliography{eccv2022submission}

\clearpage
\section{Supplementary}
\vspace{-10mm}
\begin{table} [ht!]
\caption{Setting of Early Specialization when ViT-B/16 as visual backbone, N*N means 2D kernel size of CNNs.}
     \label{tab:1}
     \centering
       \begin{tabular}{ccccc}
        \toprule
        Module& Stride&Dim& Resolution\\
        \midrule
        3*3 Conv&2 & 2$\rightarrow$48&224$\rightarrow$112\\
        Residual 3*3 Conv& 2&48$\rightarrow$96&112$\rightarrow$56\\
        Residual 3*3 Conv& 2&96$\rightarrow$192&56$\rightarrow$28\\
        Residual 3*3 Conv& 2&192$\rightarrow$384&28$\rightarrow$14\\
        Residual 3*3 Conv& 1&384$\rightarrow$768&14$\rightarrow$14\\
        1*1 Conv& 1&768$\rightarrow$768&14$\rightarrow$14\\
        \midrule
        Total \# Parameters&\multicolumn{3}{c}{4.5M}\\
        \bottomrule
    \end{tabular}
\end{table}

\vspace{-15mm}
\begin{table} [ht!]
\caption{Setting of Early Specialization when ViT-B/16 as visual backbone, N*N means 2D kernel size of CNNs.}
     \label{tab:2}
     \centering
         \begin{tabular}{ccccl}
            \toprule
            Parallel & Adapter & Fusion & Resol-\\
            Module & Module & Layer &ution\\
            \midrule
            3*3 Conv&8*8 DWConv&2&224$\rightarrow$112\\
            Bottleneck 3*3 Conv&4*4 DWConv&4&112$\rightarrow$56\\
            Bottleneck 3*3 Conv&2*2 DWConv&6&56$\rightarrow$28\\
            Bottleneck 3*3 Conv&1*1 DWConv&8&28$\rightarrow$14\\
            Bottleneck 3*3 Conv&1*1 DWConv&10&14$\rightarrow$14\\
            \midrule
            Total \# Parameters&\multicolumn{3}{c}{3.9M}\\
            \bottomrule
    \end{tabular}
\vspace{-10mm}
\end{table}

\subsection{Modality-Specific Auxiliary Module Configuration}
When visual backbone is ViT-B/16, we slight adjust the convolution kernels and strides in Early Specialization and Efficient Parallel Branch. The detailed configuration of those two are shown in Tab. \ref{tab:1} and Tab. \ref{tab:2}.

\begin{table*}[t]
\caption{Linear probing results on 24 datasets.}
\label{tab:3}
\fontsize{9}{9}\selectfont
\begin{center}
\begin{tabular}{lccc}
\toprule
\multirow{2}{*}{Datasets}  & CLIP & MS-CLIP-S&\multirow{2}{*}{$\Delta$}\\
  &(ViT-B32) & (B32)&\\
\midrule
Food-101&68.5&\textbf{\underline{76.4}}&$+$ 4.7\\
SUN397&62.0&\textbf{\underline{67.8}}&$+$ 5.8\\
Stanford Cars&70.7&\textbf{\underline{79.1}}&$+$ 8.4\\
FGVC Aircraft&38.6&\textbf{\underline{45.4}}&$+$ 6.8\\
Pascal Voc 2007&80.1&\textbf{\underline{83.9}}& $+$ 3.8\\
Describable Texture (dtd)&67.9&\textbf{\underline{75.1}}&$+$ 7.2\\
Oxford-IIIT Pets&69.4&\textbf{\underline{77.4}}&$+$ 8.0\\
Caltech-101&86.2&\textbf{\underline{88.9}}&$+$ 2.7\\
Oxford Flowers 102&89.2&\textbf{\underline{93.5}}&$+$ 4.3\\
MNIST&97.1&\textbf{\underline{98.1}}&$+$ 1.0\\
Facial Emotion Recognition&56.8&\textbf{\underline{57.2}}&$+$ 0.4\\
STL-10&93.8&\textbf{\underline{95}}&$+$ 1.2\\
GTSRB&\textbf{\underline{86.4}}&83.5&$-$ 2.9\\
PatchCamelyon&81.0&\textbf{\underline{81.1}}&$+$ 0.1\\
UCF101&70.8&\textbf{\underline{74.7}}&$+$ 3.9\\
CIFAR-10&\textbf{\underline{93.5}}&92.0&$-$ 1.5\\
CIFAR-100&\textbf{\underline{78.0}}&74.9&$-$ 3.1\\
Hateful Memes&50.6&\textbf{\underline{52.0}}&$+$ 1.4\\
ImageNet&59.1&\textbf{\underline{66.5}}&$+$ 7.4\\
Country211&13.8&\textbf{\underline{16.4}}&$+$ 2.6\\
EuroSAT&\textbf{\underline{95.1}}&94.7&$-$ 0.4\\
Kitti-distance&\textbf{\underline{44.4}}&37.6&$-$ 6.8\\
Rendered-SST2&56.8&\textbf{\underline{59.7}}&$+$ 2.9\\
Resisc45&83.0&\textbf{\underline{87.5}}&$+$ 4.5\\
\midrule
Avg.&70.5&\textbf{\underline{73.3}}&$+$ 2.8\\
\bottomrule
\end{tabular}
\end{center}
\end{table*}

\subsection{Detailed Linear Probing Results When Pre-trained on Laion-20M}
The results of linear probing on 24 various datasets with models pre-trained on Laion-20M are shown in Tab. \ref{tab:3}. Our MS-CLIP-S can outperform vanilla CLIP on 19 datasets with an average improvement of 2.7\%.

\begin{table}[t]
\renewcommand\arraystretch{1}
\setlength\tabcolsep{1pt}
\caption{Zero-shot Eval. of models pre-trained on YFCC-22M and LAION-20M. B32 denotes using ViT-B/32 as visual backbone and B16 denotes using ViT-B/16 as visual backbone.}
\vspace{-2mm}
\label{tab:zeroshot_24}
\fontsize{7}{7.3}\selectfont
\begin{center}
\begin{tabular}{lccc|ccc|ccc}
\toprule
\multirow{3}{*}{Datasets} &\multicolumn{6}{c}{YFCC-22M}  &\multicolumn{3}{c}{LAION-20M}\\
  &CLIP & MS-CLIP-S&\multirow{2}{*}{$\Delta$}  &CLIP & MS-CLIP-S&\multirow{2}{*}{$\Delta$} &CLIP & MS-CLIP-S&\multirow{2}{*}{$\Delta$}\\
 &(B32) & (B32)& &(B16) & (B16)& &(B32) & (B32)&\\
\midrule
Food-101&34.4&\textbf{\underline{41.1}}&+6.7&39.8&\textbf{\underline{40.7}}&+0.9&47.1&\textbf{\underline{56.3}}&+9.2\\
SUN397&40.4&\textbf{\underline{42.1}}&+1.7&37.6&\textbf{\underline{42.7}}&+5.0&40.2&\textbf{\underline{47.5}}&+7.3\\
Stanford Cars&1.3&\textbf{\underline{1.5}}&+0.2&1.0&\textbf{\underline{1.9}}&+0.9&13.6&\textbf{\underline{16.5}}&+2.9\\
FGVC Aircraft&2.1&\textbf{\underline{2.3}}&+0.3&\textbf{\underline{2.7}}&2.5&-0.2&3.1&\textbf{\underline{4.1}}&+1\\
Pascal Voc 2007&44.6&\textbf{\underline{48.1}}& +3.5&45.1&\textbf{\underline{48.6}}& +3.5&43.8&\textbf{\underline{48.6}}& +4.8\\
Describable Texture (dtd)&13.4&\textbf{\underline{14.6}}&+1.3&14.4&\textbf{\underline{19.5}}&+5.1&26.7&\textbf{\underline{31.4}}&+4.7\\
Oxford-IIIT Pets&\textbf{\underline{11.9}}&8.7&-3.2&11.2&\textbf{\underline{11.3}}&+0.1&50.6&\textbf{\underline{61.4}}&+1.0\\
Caltech-101&\textbf{\underline{21.7}}&19.3&-2.4&21.1&\textbf{\underline{22.9}}&+1.8&27.2&\textbf{\underline{28.7}}&+1.5\\
Oxford Flowers 102&35.4&\textbf{\underline{40.6}}&+5.1&38.5&\textbf{\underline{40.8}}&+2.3&33&\textbf{\underline{36.5}}&+3.5\\
MNIST&9.9&\textbf{\underline{10.0}}&+0.1&9.7&\textbf{\underline{10.4}}&+0.7&17.6&\textbf{\underline{25.6}}&+8\\
Facial Emotion Recognition&16.8&\textbf{\underline{19.8}}&+3.0&\textbf{\underline{17.1}}&12.4&-4.6&19.6&\textbf{\underline{23.4}}&+3.8\\
STL-10&\textbf{\underline{89.9}}&87.4&-2.5&86.8&\textbf{\underline{91.8}}&+5.0&88.4&\textbf{\underline{90}}&+1.6\\
GTSRB&7.6&\textbf{\underline{9.0}}&+1.4&4.8&\textbf{\underline{11.8}}&+7.0&\textbf{\underline{22.6}}&15.3&-7.3\\
PatchCamelyon&\textbf{\underline{50.9}} &50.0&-0.9&48.0 &\textbf{\underline{53.9}}&+5.9&\textbf{\underline{52.3}} &50.4&-1.9\\
UCF101&\textbf{\underline{32.4}} &30.4&-2.1&33.5&\textbf{\underline{34.4}} &+0.9&39&\textbf{\underline{41.8}} &+2.8\\
CIFAR-10&\textbf{\underline{79.4}}&70.2 &-9.1&\textbf{\underline{80.2}}&73.0 &-7.2&\textbf{\underline{85.1}}&81.7&-3.4\\
CIFAR-100&4.6&\textbf{\underline{4.8}} &+0.2&\textbf{\underline{4.3}}&3.1 &-1.2&\textbf{\underline{6.9}}&5.2&-1.7\\
Hateful Memes&\textbf{\underline{49.6}}&48.7 &-0.9&49.7&\textbf{\underline{52.8}}&+3.1&\textbf{\underline{53.5}}&50.8&-2.7\\
ImageNet&32.2&\textbf{\underline{36.7}}&+4.5&36.9&\textbf{\underline{39}}&+4.7&35.5&\textbf{\underline{40.2}}&+4.7\\
Country211&1.7&\textbf{\underline{2.2}}&+0.4&2.0&\textbf{\underline{2.1}}&+0.1&5.6&\textbf{\underline{7}}&+1.4\\
EuroSAT&\textbf{\underline{16.7}} &6.6&-10.1&6.1&\textbf{\underline{14.8}}&+8.7&5.6&\textbf{\underline{5.8}}&+0.2\\
Kitti-distance&13.2&\textbf{\underline{33.9}}&+20.7&19.3&\textbf{\underline{38.0}}&+18.7&\textbf{\underline{31.6}}&27.8&-3.8\\
Rendered-SST2&\textbf{\underline{51.7}} &49.9&-1.8&49.9&\textbf{\underline{50.2}}&+0.3&47.9&\textbf{\underline{50.5}}&+2.6\\
Resisc45&\textbf{\underline{24.4}} &21.2&-3.2&\textbf{\underline{29.8}}&28.4&-1.5&35.3&\textbf{\underline{37.7}}&+2.4\\
\midrule
\# \textit{Win}&10&\textbf{\underline{14}}&+4&5&\textbf{\underline{19}}&+14 &6&\textbf{\underline{18}}&+12\\
\textit{Avg.}&28.5&\textbf{\underline{29.1}}&+0.6&28.7&\textbf{\underline{31.1}}&+2.4&34.6&\textbf{\underline{36.8}}&+2.2\\
\bottomrule
\end{tabular}
\end{center}
\end{table}

\subsection{Zero-shot Evaluation on 24 datasets}
We further conduct zero-shot evaluation on all 24 datasets following the same configuration in CLIP. The complete result is shown in Tab. \ref{tab:zeroshot_24}. Our MS-CLIP-S consistently outperforms CLIP in different pre-training datasets and backbone models. When pre-trained on LAION-20M, our MS-CLIP-S outperforms CLIP on 18 out of 24 datasets with an average gain of 2.2\%. When pre-trained on YFCC-22M with ViT-B/16 as backbone, the average gain is 2.4\% with outperforming on 19 out of 24 datasets. However, when pre-trained on YFCC-22M with ViT-B/32 as backbone, the overall improvement is not that significant. We hypothesize that because of a weaker baseline, the performances in many datasets are very low and the numerical fluctuation influence a lot.  

\subsection{More Ablations}
\subsubsection{Ablation on Sharing Attention and FFN individually}

\begin{table*}[ht!]
\caption{Experimental results of sharing Attn. and FFN individually in Transformer layer.  LN1 denotes the LN before Attn. LN2 denotes the LN before FFN.}
\label{tab:share_attn_fn}
\begin{center}
\begin{tabular}{ccccc}
\toprule
Text&\multirow{2}{*}{\# Params}&Shared &Non-Shared &IN Zero-shot \\
Width& &Module& Module&Acc(\%) \\
\midrule
768&126M &Attn, FFN    & LN1, LN2 & \textbf{32.99}\\
768&154M & FFN    & Attn, LN1, LN2 & 30.40\\
768&182M &Attn    & FFN, LN1, LN2 & 26.12\\
\bottomrule
\end{tabular}
\end{center}
\end{table*}
We further conduct experiments where either FFN or Attn is shared while others are modality-specific. As in Tab. \ref{tab:share_attn_fn}, we found that still sharing both gives better result than individual sharing. We infer that it's probably because the attention modules' output is input into FFN modules, which makes them strongly coupled.

\begin{table*}[ht!]
\caption{Ablation on whether using DWConv in adapters.}
\label{tab:dwconv}
\begin{center}
\begin{tabular}{lcc}
\toprule
Model&\# Params &IN Zero-shot Acc(\%) \\
\midrule
MS-CLIP-S&132M  &\textbf{36.66}\\
 $\cdots$ w/o DWConv&131M  &33.94\\
\bottomrule
\end{tabular}
\end{center}
\end{table*}

\subsubsection{Ablation on Depth-Wise Conv in adapters}
The Depth-Wise Conv (DWConv) can gather spatial context features with 2D kernels and resize image feature map, while FFN/BottleneckFFN is applied point-wise without context. To verify the importance of spatial context, we replace DWConv with average pooling + FFN (average pooling's kernel size, stride, padding are same as DWConv) which performs worse than DWConv by 2.7\% in IN ZS accuracy, as shown in Tab. \ref{tab:dwconv}.

\end{document}